\pdfoutput=1
\documentclass[runningheads]{llncs}
\usepackage{authblk}
\usepackage{array}
\usepackage[backend=biber]{biblatex}
\usepackage{adjustbox}
\usepackage[T1]{fontenc}
\usepackage{algorithm}
\usepackage{algorithmic}
\usepackage{booktabs}
\usepackage{multirow}
\usepackage{amsmath,amssymb,amsfonts}
\usepackage{algorithmic}
\usepackage{graphicx}
\usepackage{textcomp}
\usepackage{xcolor}
\usepackage{booktabs}
\usepackage{tabularx} 
\usepackage{array}
\usepackage{float}
\usepackage{graphicx}
\begin{document}
\title{Federated Latent Factor Model for Bias-Aware Recommendation with Privacy-Preserving}
\titlerunning{\textbf{FBALF}}



\author{
  Junxiang Gao\inst{1} \and
  Yixin Ran\inst{2} \and
  Jia Chen\inst{3}
}

\institute{
  College of Computer and Information Science, Southwest University, Chongqing 400715, China\\
  \email{gjx\_wy@163.com}
  \and
  School of Engineering, University of Western Australia, Australia\\
  \email{ranyixin920@gmail.com}
  \and
  School of Cyber Science and Technology, Beihang University, Beijing 100191, China\\
  \email{chenjia@buaa.edu.cn}
}

\authorrunning{Gao et al.}

\maketitle              
\begin{abstract}
A recommender system (RS) aims to provide users with personalized item recommendations, enhancing their overall experience. Traditional RSs collect and process all user data on a central server. However, this centralized approach raises significant privacy concerns, as it increases the risk of data breaches and privacy leakages, which are becoming increasingly unacceptable to privacy-sensitive users. To address these privacy challenges, federated learning has been integrated into RSs, ensuring that user data remains secure. In centralized RSs, the issue of rating bias is effectively addressed by jointly analyzing all users' raw interaction data. However, this becomes a significant challenge in federated RSs, as raw data is no longer accessible due to privacy-preserving constraints. To overcome this problem, we propose a Federated Bias-Aware Latent Factor (FBALF) model. In FBALF, training bias is explicitly incorporated into every local model's loss function, allowing for the effective elimination of rating bias without compromising data privacy. Extensive experiments conducted on three real-world datasets demonstrate that FBALF achieves significantly higher recommendation accuracy compared to other state-of-the-art federated RSs.

\end{abstract}
\section{Introduction}
Recommender system (RS) enhance online services by providing personalized recommendations, improving user experiences, and driving business growth \cite{b1,b2,b3,b4,b5,b66,b67,b80}. Traditionally, RS rely on centralized data storage, where user-item interactions are collected on a central server to train models \cite{b6,b7,b8,b9}. However, this approach raises privacy concerns due to the sensitive nature of user data and the risk of privacy breaches \cite{b50}, further complicated by regulations such as GDPR \cite{b10}, which restrict data collection \cite{b11,b12,b13}. Federated learning offers a privacy-preserving method, enabling federated RS where raw data remains local, and only processed information (e.g., model parameters) is shared with a central server to train a global model \cite{b14,b15}.

\begin{figure}[t]
\centering
\includegraphics[scale=0.45]{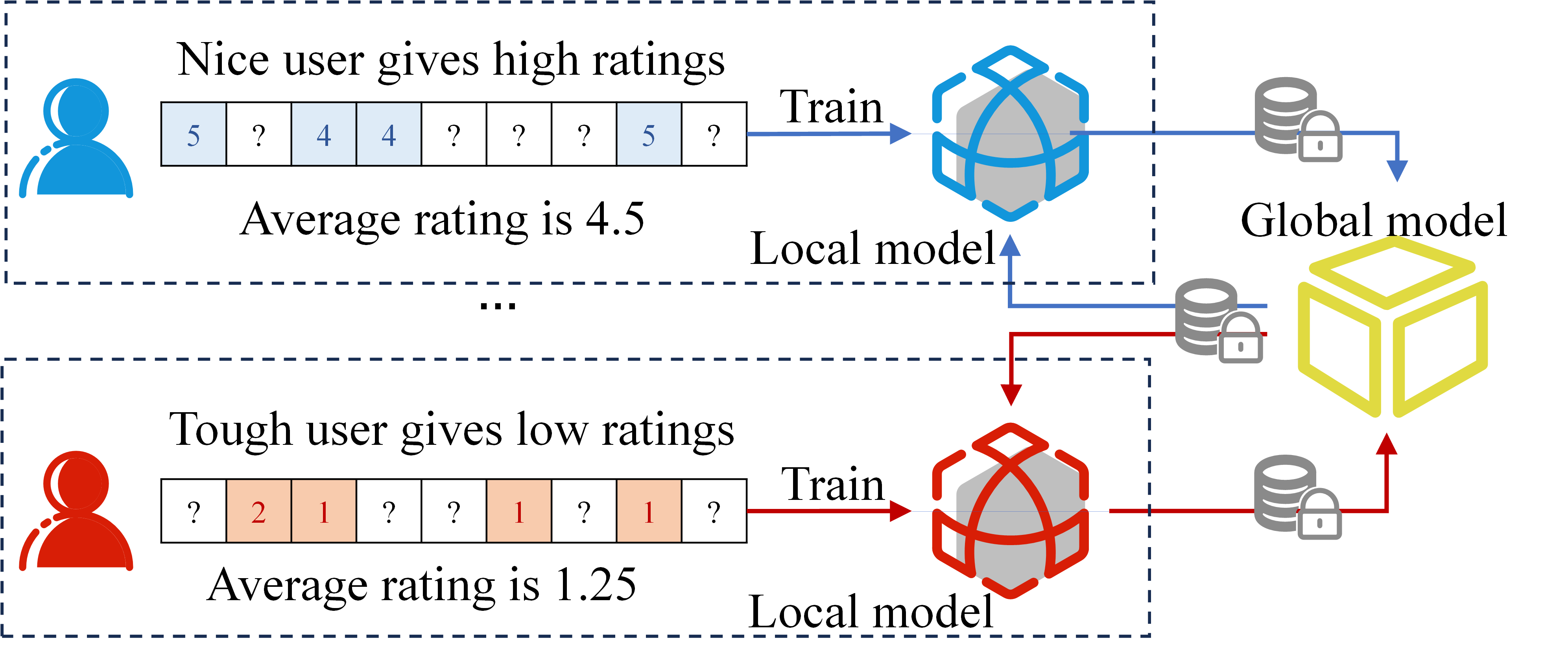}
\caption{An example illustrating rating bias in a federated RS.}
\label{fig}
\end{figure}

In real-world recommendation scenarios, a critical challenge is individual bias, where users or items consistently differ in their rating scales \cite{b16,b17,b18,b19}. For example, as shown in Fig. 1, some users habitually give higher ratings, often referred to as nice users, while others tend to provide lower ratings, known as tough users. Such systematic rating tendencies can introduce bias, distorting recommendations and reducing fairness. Traditional RSs address this issue by aggregating and analyzing raw interaction data from all users, allowing for the detection and correction of individual bias \cite{b20,b21}. However, these solutions are not applicable in federated RSs, where user data remains decentralized and inaccessible to a central server. The unavailability of raw data hinders direct bias correction, presenting considerable challenges and requiring innovative solutions in federated recommendation scenarios.

To address the issues mentioned above, we propose the Federated Bias-Aware Latent Factor (FBALF) model. Training bias is incorporated into the local model's loss function to eliminate the rating bias. Additionally, we employ a noise strategy to further protect each user's private interaction data, a consideration that is often neglect in related models. The main contributions of this paper can be summarized as follows:
\begin{enumerate}
    \item We propose the FBALF model, which effectively captures user and item representations in a federated setting, leading to improved recommendation performance.
    \item We introduce training bias in a federated setting to mitigate the impact of user and item rating bias on predictions, while employing hybrid filling to protect user preferences from leakage.
    \item Extensive empirical experiments are conducted on three datasets derived from industrial applications to evaluate the proposed model and other state-of-the-art models.
\end{enumerate}

\section{Related Work}
In the rating prediction task of RSs, the data source is a highly sparse user-item rating matrix \cite{b22,b23,b24,b25,b26,b51,b52,b53,b54,b63}. While deep learning-based RSs have achieved significant success \cite{b1,b27,b55,b56,b79}, they suffer from high computational overhead because they process the entire sparse matrix, including missing entries \cite{b28,b29,b68}. In contrast, latent factor models are computationally efficient, as they are trained only on observed ratings \cite{b57,b58,b59,b60,b62}. This efficiency, combined with their accuracy and scalability, makes latent factor models a widely adopted solution for rating prediction tasks \cite{b31,b32,b61}. 

In the past decade, many latent factor model-based rating predictors have been proposed from different perspectives, including a basic latent factor-based model \cite{b33}, a generalized non-negative and momentum-based model \cite{b34}, a extended linear bias-based model \cite{b20}, a generalized non-negative model \cite{b35}, a multi-norms-based model \cite{b23}, a multi-metric latent feature based model \cite{b36}, and a prediction-sampling-based model \cite{b16}. However, all of the above models are centralized, which poses a risk of privacy leakage. In this paper, we adopt the latent factor model as the base model to develop a federated rating predictor, transferring the merits of high accuracy and computational efficiency from latent factor models to federated RS.

RSs keep users' raw data local, sharing only intermediate results like model parameters or updating gradients. Recently, several matrix factorization based federated models have been proposed, including federated collaborative filtering \cite{b13}, decentralized matrix factorization \cite{b11}, meta matrix factorization \cite{b37}, and robust regularized federated matrix factorization \cite{b38}. However, privacy risks persist when transferring intermediate results. Differential privacy (DP) adds noise to anonymize data but reduces accuracy \cite{b39}. Although these models protect raw rating data, users' rating behaviors (rated items) remain identifiable since rated items' intermediate results are shared \cite{b40}. To address this, Lin et al. introduced hybrid filling to generate virtual ratings for unrated items. Liang et al. later improved this by applying a denoising strategy to mitigate the adverse effects of virtual ratings \cite{b41}. However, these models neglect the rating bias associated with both users and items.

In summary, as illustrated in Fig. 1, real federated recommendation scenarios involve various users and items, which inevitably leads to rating bias. The aforementioned federated models do not take this issue into account. In contrast, our FBALF considers how to eliminate the impact of bias in a federated setting. Therefore, our FBALF achieves better prediction accuracy.
 
\section{PRELIMINARIES}
\subsection{Symbols and Notations}
Table I provides a detailed description of the symbols and notations used in the following sections.

\begin{table}[t]
    \caption{Symbols and Notations}
    \centering
    \resizebox{0.8\textwidth}{!}{
    \begin{tabular}{>{\centering}m{2cm}>{\centering\arraybackslash}m{8cm}}
        \toprule[1pt]
        Symbol & Explanation \\
        \midrule 
        $U, I$ & The set of users and the set of targeted items. \\
        $u, i$ & A user from $U$ and an item from $I$. \\
        $R$ & The user-item rating matrix, where $|U|$ is the number of users and $|I|$ is the number of items, representing a sparse rating matrix. \\
        $R_K$ & The set of known ratings from users in the matrix. \\
        $r_{u,i}, \hat{r}_{u,i}$ & The actual rating given by user $u \in U$ on item $i \in I$ and the predicted rating for the same. \\
        $I_u, \hat{I}_u$ & The set of items rated by user $u$ and the synthetic ratings generated for user $u$. \\
        $f$ & The dimensionality of the latent factor space. \\
        $C, S$ & The latent factor matrices for users and items, where $C$ has $|U|$ rows and $f$ columns, and $S$ has $|I|$ rows and $f$ columns. \\
        $a, b$ & The bias vectors for users and items, respectively. \\
        $T$ & The maximum number of iterations for the training process. \\
        $o_u$ & An indicator variable indicating whether item $i$ has been rated by user $u$. \\
        $\rho$ & A parameter controlling the number of sampled unrated items. \\
        $r_{u,i}^{\prime}$ & The synthetic rating generated using the hybrid filling strategy. \\
        $T_{\mathrm{HF}}, T_{\mathrm{local}}$ & The threshold for the iteration that determines the approach used for generating synthetic ratings, and the number of local training iterations for user $u$, respectively. \\
        $\lambda$ & The Tikhonov regularization parameter. \\
        $\eta$ & The learning rate. \\
        $\Gamma$ & The testing set. \\
        \bottomrule [1pt]
    \end{tabular}
    }
\end{table}

\subsection{Centralized Latent Factor Model}
Suppose there is a user set \( U \) and an item set \( I \), where each user \( u \) (\( u \in U \)) has rated a subset of items, denoted as \( I_u \) (\( I_u \subseteq I \)). Let \( R \) be a matrix with \( |U| \) rows and \( |I| \) columns, where each entry \( r_{u,i} \) at the \( u \)-th row and \( i \)-th column records the observed rating of item \( i \) (\( i \in I_u \)) rated by user \( u \). 
Since users typically rate only a small portion of items, \( R \) is inherently sparse \cite{b74,b75}. We refer to \( R \) as a centralized sparse user-item rating matrix.

Given \( R \), a centralized latent factor model employs two latent factor matrices \( C \) and \( S \) to make \( R \)'s rank-\( f \) approximation by minimizing the sum of errors between \( R \) and its approximation only on \( R \)'s observed ratings. With \( C \) and \( S \), the approximation of \( R \) is obtained by:$\hat{R} = C S^T$. To evaluate the approximation, a metric is needed to measure the error. According to \cite{b42,b43,b64,b65,b76}, the Euclidean distance is a common metric.  
Then, the objective function of a centralized latent factor model can be formulated as follows:

\begin{equation}
\begin{aligned}
&\min_{C, S}  \frac{1}{2} \sum \left\| R - CS^{\mathsf{T}} \right\|_F^2, \\
\text{s.t. } & r_{u,i} \in \left\{ r_{u,i} \mid u \in \mathcal{U}, i \in I_u \right\}
\end{aligned}
\end{equation}
where \( \| \cdot \|_F \) denotes the Frobenius norm of a matrix. Additionally, regularization plays a crucial role in preventing overfitting \cite{b44,b66,b69,b77,b78}. By incorporating Tikhonov regularization into a centralized latent factor model, equation (1) is modified as follows:

\begin{equation}
\begin{aligned}
\mathop {\min }\limits_{C,S{\rm{ }}} \frac{1}{2}\sum &{\left\| {R - C{S^{\rm T}}} \right\|_F^2}  + \frac{\lambda }{2}\left( {\left\| C \right\|_F^2 + \left\| S \right\|_F^2} \right)\\
&s.t.{\rm{  }}{r_{u,i}} \in \{ {\left. {{r_{u,i}}} \right|u \in {\cal U},{\rm{ }}i \in {I_u}}\}
\end{aligned}
\end{equation}
where \( \lambda \) is a hyperparameter that controls the regularization intensity and (2) is minimized using stochastic gradient descent (SGD) to obtain the updating rules for \( C \) and \( S \).

\section{Proposed FBALF}
\subsection{Problem: Federated Rating Prediction}
Consider a user set \( U \) and an item set \( I \), where each user \( u \) (\( u \in U \)) has rated a subset \( I_u \subseteq I \). The problem of federated rating prediction is to predict the potential ratings of each user \( u \) for their unrated items \( I \setminus I_u \), while ensuring that the raw rating data of each user \( u \) (i.e., \( \{ r_{u,i} \mid \forall i \in I_u \} \)) and their rating behaviors (i.e., \( I_u \)) remain private and are not shared with other users or any third-party organization (e.g., a global server).

\subsection{Objective Formulation}
To explain the construction of a FBALF model, we first describe the objective function for the \( u \)-th user, where \( u \in \{1, 2, \dots, |U|\} \). Based on (2), the objective function for the \( u \)-th user can be expressed as follows:

\begin{equation}
\begin{aligned}
\begin{array}{l}
\mathop {\min }\limits_{{c_u},{\rm{ }}S} {\varepsilon _u}=
\mathop {\min }\limits_{{c_u},{\rm{ S}}} \frac{1}{2}\sum\limits_{i \in {I_u}} {{{\left( {{r_{u,i}} - \sum\limits_{k = 1}^f {{c_{u,k}}{s_{i,k}}} } \right)}^2}}
\\
+ {\rm{ }}\frac{\lambda }{2}\sum\limits_{i \in {I_u}} {\left( {\sum\limits_{k = 1}^f {{{\left( {{c_{u,k}}} \right)}^2} + \sum\limits_{k = 1}^f {{{\left( {{s_{i,k}}} \right)}^2}} } } \right)}
\end{array}
\end{aligned}
\end{equation}
where \( c_{u,k} \) and \( s_{i,k} \) are the \( k \)-th entries of \( c_u \) and \( s_i \), respectively.

\subsubsection{Training Bias}
From (2), a latent factor model primarily captures the interactions between users and items. However, a significant portion of the variance in rating values is due to bias or intercepts associated with users or items, which are independent of interactions. For example, different users often have varying rating tendencies: some users are generous, tending to give higher ratings, while others are strict, tending to give lower ratings. To improve the accuracy and personalization of recommendations, it is essential to incorporate bias in each user’s ratings. Similarly, items also exhibit diverse rating patterns, with some items consistently receiving higher or lower ratings than others.

Therefore, relying solely on latent factors for rating prediction is insufficient. Building upon equation (3), the objective function with training bias can be expressed as follows:

\begin{equation}
\begin{aligned}
\begin{array}{l}
\mathop {\min }\limits_{{c_u},S,{a_u},b} {\varepsilon _u} = \\
\mathop {\min }\limits_{{c_u},S,{a_u},b} \frac{1}{2}\sum\limits_{i \in {I_u}} {{{\left( {{r_{u,i}} - {a_u} - {b_i} - \sum\limits_{k = 1}^f {{c_{u,k}}{s_{i,k}}} } \right)}^2}} \\
\quad\quad\quad + \frac{\lambda }{2}\sum\limits_{i \in {I_u}} {\left( {\sum\limits_{k = 1}^f {{{\left( {{c_{u,k}}} \right)}^2} + \sum\limits_{k = 1}^f {{{\left( {{s_{i,k}}} \right)}^2}} } } \right)}
\end{array}
\end{aligned}
\end{equation}
Where \( a_u \) and \( b_i \) represent the observed deviations for user \( u \) and item \( i \), respectively. Additionally, regularizing the bias terms is crucial for improving the model's generalization capability. By incorporating Tikhonov regularization into the bias, let \( \hat{r}_{u,i} = a_u + b_i + \sum_{k=1}^f c_{u,k} s_{i,k} \), the objective function can be reformulated as:

\begin{equation}
\begin{aligned}
\begin{array}{l}
\mathop {\min }\limits_{{c_u},S,{a_u},b} {\varepsilon _u} =
\mathop {\min }\limits_{{c_u},S,{a_u},b} \frac{1}{2}\sum\limits_{i \in {I_u}} {{{\left( {{r_{u,i}} - {{\hat r}_{u,i}}} \right)}^2}} \\
+\frac{\lambda }{2}\sum\limits_{i \in {I_u}} {\left( {\sum\limits_{k = 1}^f {{{\left( {{c_{u,k}}} \right)}^2} + \sum\limits_{k = 1}^f {{{\left( {{s_{i,k}}} \right)}^2}} }  + {{\left( {{a_u}} \right)}^2} + {{\left( {{b_i}} \right)}^2}} \right)}
\end{array}
\end{aligned}
\end{equation}

\subsubsection{Hybrid Filling}
In federated learning, safeguarding the privacy of local user data, such as click preferences, is of utmost importance. In this paper, we use an effective hybrid filling strategy to protect users' privacy.

As discussed before, each user only rates a small subset \( I_u \) of items from \( I \), resulting in a lack of specific rating data for most items. To address this, we assign synthetic ratings for some missing data. First, we define the synthetic set of items \( \hat{I}_u \subseteq I \setminus I_u \) for which ratings need to be assigned, with \( |\hat{I}_u| = \rho |I_u| \), where \( \rho \) is a parameter that controls the number of unrated items to be sampled. Then, for a user \( u \in U \) and an item \( i \in \hat{I}_u \), a synthetic rating \( r_{u,i}' \) is computed as follows:
\begin{equation}
\begin{aligned}
\hat{r}_{u,i}' = 
\begin{cases} 
\frac{\sum_{i \in I_u} r_{u,i}}{|I_u|}, & t \leq T_{HF} \\ 
\ \ \ \ \hat{r}_{u,i}, & t > T_{HF} 
\end{cases}
\end{aligned}
\end{equation}
where \( t \) represents the \( t \)-th training iteration, \(T_{HF} \) is the iteration threshold, and different methods for generating synthetic ratings are applied at different stages of the training process.

This synthetic rating set effectively addresses privacy concerns in federated learning. First, it becomes difficult for the server to distinguish which ratings are truly provided by a user in the combined rating set \( I_u \cup \hat{I}_u \), thereby safeguarding the user's preferences. Second, the parameter \( \rho \) controls the size of the augmented rating set, introducing only minimal additional computational and communication overhead.

To differentiate between the two types of items in the combined set \( I_u \cup \hat{I}_u \), we define \( o_{u,i} \in \{0, 1\} \) as an indicator variable:

\begin{equation}
\begin{aligned}
{o_{u,i}} = \left\{ \begin{array}{l}
1,\quad \quad \text{if} \  i \in {I_u}\\
0,\quad \quad \text{if} \ i \in {{\hat I}_u}
\end{array} \right.
\end{aligned}
\end{equation}
then, the objective function for user \( u \) can then be redefined as:
\begin{equation}
\begin{aligned}
\begin{array}{l}
{\rm{     }}\mathop {\min }\limits_{{c_u},S,{a_u},b} {\varepsilon _u} = \\
{\rm{     }}\mathop {\min }\limits_{{c_u},S,{a_u},b} \frac{1}{2}\sum\limits_{i \in {I_u} \cup {{\hat I}_u}} {{{\left( {{o_{u,i}}{r_{u,i}} + \left( {1 - {o_{u,i}}} \right){{r'}_{u,i}} - {{\hat r}_{u,i}}} \right)}^2}} \\
{\rm{              +  }}\frac{\lambda }{2}\sum\limits_{i \in {I_u} \cup {{\hat I}_u}} {\left( {\sum\limits_{k = 1}^f {{{\left( {{c_{u,k}}} \right)}^2} + \sum\limits_{k = 1}^f {{{\left( {{s_{i,k}}} \right)}^2}} }  + {{\left( {{a_u}} \right)}^2} + {{\left( {{b_i}} \right)}^2}} \right)} .
\end{array}
\end{aligned}
\end{equation}

\subsection{Model Optimization}
For FBALF, the main idea is to determine the optimal latent factors \( C \) and \( S \), as well as the training bias \( a \) and \( b \). While \( S \) and \( b \) are stored on the central server, \( C \) and \( a \) are stored on the local users. We will describe the model training process for local users and the central server separately.
\subsubsection{Training in The Local User}
For a given local user \( u \), the user needs to download the randomly initialized latent factors \( S \) and training bias \( b \) from the server. The user does not require access to the latent factors or training bias of other users to compute the local gradients of its own latent factors \( c_u \) and training bias \( a_u \). Moreover, utilizing the hybrid filling strategy, we generate synthetic ratings for a randomly sampled subset \( \hat{I}_u \) to augment the original rating data. With the augmented rating set \( I_u \cup \hat{I}_u \), and by considering the loss for a single item \( i \in I_u \cup \hat{I}_u \), we reformulate (8) into a single-element-oriented form:

\begin{equation}
\begin{aligned}
\begin{array}{l}
\mathop {\min }\limits_{{c_u},{\rm{ }}{a_u}} {\varepsilon _{u,i}} = \mathop {\min }\limits_{{c_u},{\rm{ }}{a_u}} \frac{1}{2}{\left( {{o_{u,i}}{r_{u,i}} + \left( {1 - {o_{u,i}}} \right){{r'}_{u,i}} - {{\hat r}_{u,i}}} \right)^2}\\
\quad \quad {\rm{ +  }}\frac{\lambda }{2}\left( {{{\left( {{a_u}} \right)}^2} + {{\left( {{b_i}} \right)}^2} + \sum\limits_{k = 1}^f {{{\left( {{c_{u,k}}} \right)}^2} + \sum\limits_{k = 1}^f {{{\left( {{s_{i,k}}} \right)}^2}} } } \right)
\end{array}
\end{aligned}
\end{equation}

Then, the optimization of \( c_u \) and \( a_u \) can be achieved by SGD. Denote $\Delta_{u,i}^{t-1} = o_{u,i} r_{u,i} + (1 - o_{u,i}) r_{u,i}^{\prime} - \hat{r}_{u,i}$, the gradients \( \nabla c_{u,k}^{t-1} \) and \( \nabla a_u^{t-1} \) can be computed as:

\begin{equation}
\begin{aligned}
\left\{
\begin{array}{l}
\nabla c_{u,k}^{t - 1} =  - \eta s_{i,k}^{t - 1}\Delta _{u,i}^{t - 1} + \eta \lambda c_{u,k}^{t - 1}\\
\nabla a_u^{t - 1} =  - \eta \Delta _{u,i}^{t - 1} + \eta \lambda a_u^{t - 1}
\end{array}
\right.
\end{aligned}
\end{equation}
where \( s_{i,k}^{t-1} \), \( c_{u,k}^{t-1} \), and \( a_u^{t-1} \) are the states of \( s_{i,k} \), \( c_{u,k} \) and \( a_u \) at the \( (t-1) \)-th iteration. 

Then, at the \( t \)-th iteration, (9) can be minimized by:

\begin{equation}
\begin{aligned}
\left\{
\begin{array}{l}
\text{for } k = 1 \sim f, \ \  c_{u,k}^{t} \leftarrow c_{u,k}^{t - 1} - \eta \nabla c_{u,k}^{t - 1} \\
\hspace{1.7cm} a_u^{t} \leftarrow a_u^{t - 1} - \eta \nabla a_u^{t - 1}
\end{array}
\right.
\end{aligned}
\end{equation}
where \( \eta \) is the learning rate.

The gradients \( \nabla s_{i,k}^{t-1} \) and \( \nabla b_u^{t-1} \) can be obtained similarly to (10), but instead of being used for local updates, they are uploaded to the central server subsequently.

\subsubsection{Training in The Central Server}
Similar to the user side, the single-element-oriented objective function is defined as follows:

\begin{equation}
\begin{aligned}
\begin{array}{l}
\mathop {\min }\limits_{{s_i},{\rm{ }}{b_i}} {\varepsilon _{u,i}} = \mathop {\min }\limits_{{s_i},{\rm{ }}{b_i}} \frac{1}{2}{\left( {{o_{u,i}}{r_{u,i}} + \left( {1 - {o_{u,i}}} \right){{r'}_{u,i}} - {{\hat r}_{u,i}}} \right)^2}\\
\ \ \ \ {\rm{ +  }}\frac{\lambda }{2}\left( {{{\left( {{a_u}} \right)}^2} + {{\left( {{b_i}} \right)}^2} + \sum\limits_{k = 1}^f {{{\left( {{c_{u,k}}} \right)}^2} + \sum\limits_{k = 1}^f {{{\left( {{s_{i,k}}} \right)}^2}} } } \right)
\end{array}
\end{aligned}
\end{equation}

Once the local users complete their updates and upload the gradients \( \nabla s_{i,k}^{t-1}(u) \) and \( \nabla b_{i}^{t-1}(u) \), the server utilizes these gradients for updating.  
Specifically, with \( \nabla s_{i,k}^{t-1}(u) \) and \( \nabla b_{i}^{t-1}(u) \) uploaded by user $u$, (12) can be minimized by:

\begin{equation}
\begin{aligned}
\left\{
\begin{array}{l}
\text{for } k = 1 \sim f, \ \  s_{i,k}^{t} \leftarrow s_{i,k}^{t - 1} - \eta \nabla s_{i,k}^{t - 1}(u) \\
\hspace{1.7cm}  b_{i}^{t} \leftarrow b_{i}^{t - 1} - \eta \nabla b_{i}^{t - 1}(u)
\end{array}
\right.
\end{aligned}
\end{equation}

\section{EXPERIMENTS}
\subsection{Experimental settings}
\subsubsection{Dataset} We use three frequently used sparse datasets ML1M , Yahoo , and Hetrec-ML to evaluate FBALF, Table II summarizes their details.
\begin{table}[t]
    \caption{Properties of All the Datasets}
    \centering
\begin{tabularx}{\textwidth}{c X X X X X}
        \toprule
        No. & Name & \( |U| \) & \( |I| \) & \( |R_K| \) & Density \\
        \midrule
        D1 & ML1M       & 6,040   & 3,706    & 1,000,209 & 4.47\% \\
        D2 & Yahoo      & 15,400  & 1,000    & 365,704   & 3.37\% \\
        D3 & Hetrec-ML  & 2,113   & 10,109   & 855,598   & 4.01\% \\
        \bottomrule
    \end{tabularx}
\end{table}

\subsubsection{Evaluation Metrics}
Predicting missing data is a critical challenge in assessing the representational capability of a user-item rating matrix. To measure the accuracy of prediction, mean absolute error (MAE) and root mean squared error (RMSE) are widely used as evaluation metrics \cite{b45,b46,b47,b48,b72,b73}. These evaluation metrics are defined as follows:
\[{\rm{ MAE}} = {{\left( {\sum\limits_{{r_{u,i}} \in \Gamma } {\left| {{r_{u,i}} - {{\hat r}_{u,i}}} \right|} } \right)} \mathord{\left/
 {\vphantom {{\left( {\sum\limits_{{h_{u,i}} \in \Gamma } {\left| {{r_{u,i}} - {{\hat r}_{u,i}}} \right|} } \right)} {\left| \Gamma  \right|}}} \right.
 \kern-\nulldelimiterspace} {\left| \Gamma  \right|}}\]
\[{\rm{RMSE}} = \sqrt {{\left( {{{\sum\limits_{{r_{u,i}} \in \Gamma } {\left( {{r_{u,i}} - {{\hat r}_{u,i}}} \right)} }^2}} \right)} \mathord{\left/
 {\vphantom {{\left( {{{\sum\limits_{{r_{u,i}} \in \Gamma } {\left( {{r_{u,i}} - {{\hat r}_{u,i}}} \right)} }^2}} \right)} {\left| \Gamma  \right|}}} \right.
 \kern-\nulldelimiterspace} {\left| \Gamma  \right|}}\]
where \( \Gamma \) denotes the testing set and lower MAE and RMSE values indicate better prediction accuracy.
\subsubsection{Baselines}  
To evaluate the effectiveness of the proposed FBALF model, we compare it with five state-of-the-art models: FedMF \cite{b14}, a basic federated matrix factorization framework; FedRec \cite{b40}, a federated framework that preserves privacy with averaging and hybrid strategies; MetaMF \cite{b37}, a federated model generating item embeddings using a meta-network; FedRec++ \cite{b41}, a federated model improving privacy via noisy data and denoising clients; and RFRec \cite{b38}, an efficient and robust regularized federated recommendation model.
\begin{table*}[t]
\caption{THE COMPARISON RESULTS OF RATING ACCURACY, INCLUDING THE LOSS/WIN COUNTS, WILCOXON SIGNED-RANKS TEST, AND FRIEDMAN TEST}
\centering
\renewcommand{\arraystretch}{1.25}
\setlength{\tabcolsep}{14pt}
\resizebox{\textwidth}{!}{
\begin{tabular}{c<{\centering}c<{\centering}c<{\centering}c<{\centering}c<{\centering}c<{\centering}c<{\centering}c<{\centering}}
\hline
Dataset                               & Metric   & FedMF             & FedRec            & MetaMF            & FedRec++          & RFRec             & FBALF           \\ \hline
\multirow{2}{*}{D1}                   & MAE      & 0.6981            & 0.6867            & 0.6905            & 0.7096            & 0.6893            & \textbf{0.6639} \\
                                      & RMSE     & 0.8919            & 0.8753            & 0.8769            & 0.9105            & 0.8728            & \textbf{0.8442} \\ \cline{2-8} 
\multirow{2}{*}{D2}                   & MAE      & 0.9683            & 0.9394            & 0.9680            & 0.9062            & 1.0314            & \textbf{0.8817} \\
                                      & RMSE     & 1.2387            & 1.2244            & 1.2273            & 1.1794            & 1.2751            & \textbf{1.1607} \\ \cline{2-8} 
\multirow{2}{*}{D3}                   & MAE      & 0.5844            & 0.5906            & 0.5869            & 0.6155            & 0.7184            & \textbf{0.5713} \\
                                      & RMSE     & 0.7762            & 0.7830            & 0.7772            & 0.8149            & 0.9495            & \textbf{0.7523} \\ \hline
\multirow{3}{*}{Statistical Analysis} & loss/win & 0/6               & 0/6               & 0/6               & 0/6               & 0/6               & \textbf{Total 0/30}   \\
                                      & p-value  & \textbf{0.015625} & \textbf{0.015625} & \textbf{0.015625} & \textbf{0.015625} & \textbf{0.015625} & -               \\
                                      & F-rank   & \textbf{4.0000}   & \textbf{3.1667}   & \textbf{3.6667}   & \textbf{4.3333}   & \textbf{4.8333}   & \textbf{1.0000}  \\ \hline
\end{tabular}}
\end{table*}

\subsubsection{Experimental Designs}  
We exclude users and items with fewer than 10 ratings. For each dataset, 80\% of the ratings are used for training, while the remaining 20\% are reserved for testing. The experiments are conducted using five-fold cross-validation. We set the latent factor dimension $f = 20$ for FedMF, FedRec, FedRec++, RFRec, and our FBALF. We follow the settings recommended by the original papers and tune the hyperparameters when necessary to achieve optimal performance. For the FBALF model, we fix the threshold for hybrid filling \( T_{\text{HF}} \) and the iteration number \( T_{\text{local}} \) for locally training the user's latent factors and bias to 10. Other hyperparameters are set as follows: learning rate $\eta \in \{0.001, 0.002, 0.003\}$; Tikhonov regularization coefficient $\lambda \in \{0.02, 0.04, 0.06, 0.08, 0.1\}$; sampling hyperparameter $\rho \in \{0, 1, 2, 3\}$; and the maximum number of iterations $T = 300$.

\subsection{Performance Comparison}

To validate the effectiveness of the proposed FBALF in representation, we compare it with five state-of-the-art models. Table III summarizes the prediction accuracy of all the involved models across different datasets. For a more thorough results analysis, we performed statistical evaluations using the loss/win count, the Wilcoxon signed-ranks test, and the Friedman test \cite{b49,b70,b71}. The loss/win count measures how many times FBALF outperforms or underperforms each comparison model in terms of rating prediction accuracy (measured by MAE and RMSE) across all datasets. The Wilcoxon signed-ranks test is a non-parametric method for pairwise comparisons, used to determine whether FBALF significantly outperforms each comparison model based on the p-value. The Friedman test assesses the performance of multiple models across multiple datasets simultaneously, using the F-rank value, where a lower F-rank value indicates better prediction accuracy. The statistical results of the loss/win count, Wilcoxon signed-ranks test, and Friedman test are shown in the third-to-last, second-to-last, and last rows of Table III, respectively.

Based on Table III, we have the following notable observations:
\begin{enumerate}
    \item FBALF achieves the lowest MAE and RMSE in all cases. Specifically, the total loss/win cases situation is 0/30, meaning FBALF outperforms all comparison models (FedMF, FedRec, MetaMF, FedRec++, and RFRec) in every dataset and metric (MAE and RMSE). This demonstrates its superior performance in rating prediction accuracy.
    \item All p-values in Table III are below 0.05, indicating that FBALF significantly outperforms all comparison models in rating prediction accuracy at a 0.05 significance level on the tested datasets. 
    \item FBALF achieves the lowest F-rank value, with an F-rank of 1.0000, confirming its superior rating prediction accuracy across the tested datasets. This statistical significance further reinforces the robustness and reliability of FBALF's performance.
\end{enumerate}

\section{Conclusions}
In this paper, we propose a federated bias-aware latent factor (FBALF) model to improve recommendation accuracy while safeguarding user privacy. The model integrates training bias to enhance its performance by addressing inherent biases in user-item interactions. It operates within a federated learning framework, ensuring that user data remains decentralized, thereby protecting sensitive information. To further enhance privacy, we employ a hybrid filling strategy that generate synthetic ratings to protect users' preference privacy during the recommendation process. The experimental results demonstrate that our proposed model achieves superior recommendation accuracy compared to other state-of-the-art federated models.

\printbibliography
\end{document}